\documentclass[conference]{IEEEtran}
\usepackage{enumerate}
\usepackage{booktabs}
\usepackage{multirow}
\usepackage{algorithm}
\usepackage{algpseudocode}
\usepackage{comment}
\usepackage{amsmath,amssymb}
\usepackage{latexsym}
\usepackage{graphicx}
\usepackage{subfigure}
\usepackage{epstopdf}
\usepackage{amssymb}
\usepackage{epsfig}
\usepackage{xspace}
\usepackage[bottom]{footmisc}
\usepackage[11pt]{moresize}
\usepackage{todonotes}
\usepackage{url}
\IEEEoverridecommandlockouts
\newcommand{\mypara}[1]{
	\vspace*{0.01cm}
	\noindent\textbf{\textit{#1}}}
\newcommand{\RNum}[1]{\uppercase\expandafter{\romannumeral #1\relax}}
\DeclareMathAlphabet{\mathpzc}{OT1}{pzc}{m}{it}
\newcommand{\model}{\textbf{\texttt{HDTest}}}

\begin{document}

\title{HDTest: Differential Fuzz Testing of Brain-Inspired Hyperdimensional Computing}
\author{
	Dongning Ma$^{\ddag}$, Jianmin Guo$^{\S}$, Yu Jiang$^{\S}$, Xun Jiao$^{\ddag}$, \\
	$^{\ddag}$Villanova University,
	$^{\S}$Tsinghua University\\
	\vspace{-0.5cm}
	\thanks{Xun Jiao$^{\ddag}$ is the corresponding author}
}
\maketitle
\begin{sloppypar}

\begin{abstract}
Brain-inspired hyperdimensional computing (HDC) is an emerging computational paradigm that mimics brain cognition and leverages hyperdimensional vectors with fully distributed holographic representation and (pseudo)randomness. 
Compared to other machine learning (ML) methods such as deep neural networks (DNNs), HDC offers several advantages including high energy efficiency, low latency, and one-shot learning, making it a promising alternative candidate on a wide range of applications. However, the reliability and robustness of HDC models have not been explored yet. In this paper, we design, implement, and evaluate \model ~to test HDC model by automatically exposing unexpected or incorrect behaviors under rare inputs. The core idea of \model ~is based on guided differential fuzz testing. Guided by the distance between query hypervector and reference hypervector in HDC, \model ~continuously mutates original inputs to generate new inputs that can trigger incorrect behaviors of HDC model. Compared to traditional ML testing methods, \model ~does not need to manually label the original input. Using handwritten digit classification as an example, we show that \model ~can generate thousands of adversarial inputs with negligible perturbations that can successfully fool HDC models. On average, \model ~can generate around 400 adversarial inputs within one minute running on a commodity computer. Finally, by using the \model-generated inputs to retrain HDC models, we can strengthen the robustness of HDC models. To the best of our knowledge, this paper presents the first effort in systematically testing this emerging brain-inspired computational model. 
\end{abstract}

\section{Introduction}
Hyperdimensional computing (HDC) is an emerging computing scheme based on the working mechanism of brain that computes with deep and abstract patterns of neural activity instead of actual numbers. Compared with traditional ML algorithms such as DNN, HDC is more memory-centric, granting it advantages such as relatively smaller model size, less computation cost, and one-shot learning, making it a promising candidate in low-cost computing platforms~\cite{ge2020classification}. Recently, HDC has demonstrated promising capability on various applications such as language classification~\cite{rahimi2016robust}, vision sensing~\cite{hersche2020integrating}, brain computer interfaces~\cite{rahimi2017hyperdimensional}, gesture recognition~\cite{moin2018emg}, and DNA pattern matching~\cite{kim2020geniehd}. However, despite the growing popularity of HDC, the reliability and robustness of HDC models have not been systematically explored yet. 

The traditional approach to testing ML systems is to manually label a set of data and input them into the ML system to assess the accuracy. However, as the ML systems are scaling significantly and the input space is becoming more sophisticated, such manual approach is hardly feasible and scalable anymore. On the other side, researchers have found that by adding even invisible perturbations to original inputs, ML systems can be ``fooled'' and produce wrong predictions~\cite{szegedy2013intriguing}. 

\begin{figure}
    \centering
    \subfigure[]{
        \includegraphics[keepaspectratio, width=.14\columnwidth, page=1]{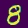}
    }
    \subfigure[]{
        \includegraphics[keepaspectratio, width=.14\columnwidth, page=1]{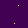}
    }
    \subfigure[]{
        \includegraphics[keepaspectratio, width=.14\columnwidth, page=1]{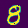}
    }
    \label{fig:case}
    \caption{An example of adversarial image of HDC by mutating some pixels in the image:(a) the original image as ``8''; (b) the pixels mutated; (c) the mutated image wrongly predicted as ``3''.}
\end{figure}

Just like DNNs, HDC can also be vulnerable to small perturbations on inputs to produce wrong classification, as shown in Fig.~\ref{fig:case}. Our attempt to enable automatic testing on HDC models faces the following challenges. First, unlike traditional ML systems such as DNNs with a well-defined mathematical formulation and relatively fixed architecture (specific layer types and network structure), HDC is less application-agnostic. The encoding of HDC is largely unique for different applications and relies on random indexing to project data onto vectors in a hyperdimensional space~\cite{rahimi2016robust}, adding difficulty to efficiently acquire adequate information to guide the testing process. As a result, adversarial generation techniques used in DNNs cannot be applied here since they rely on a set of well-defined mathematical optimization problems~\cite{goodfellow2014explaining}. Second, the standard approach to testing ML systems is to gather and manually label as much real-world test data as possible~\cite{russakovsky2015imagenet, wang2019exploring}. Google even used simulation to generate synthetic data~\cite{madrigal2017inside}. However, such manual effort is not only largely short of scalability but also unguided as it does not consider the internal structure of ML systems, making it unable to cover more than a tiny fraction of all possible corner cases. 



To address such challenges, we present \model, a highly-automated testing tool specifically designed for HDC based on differential fuzz testing. Fuzz testing is a software testing technique that strategically mutates inputs with the goal of generating faults or exceptions~\cite{liang2019deepfuzzer}. Assuming a greybox testing scenario, we leverage the unique property of HDC to guide and customize the fuzzing process. To avoid manual checking effort, we integrate fuzz testing with differential testing techniques~\cite{mckeeman1998differential},  
 making it a highly automated and scalable testing solution. 
Our contributions are as follows: 
\begin{itemize}
	\item To the best of our knowledge, we present the first effort in systematically testing HDC models. Based on differential fuzz testing, \model ~iteratively mutates inputs to generate new inputs that can trigger incorrect behaviors of HDC models. Thanks to differential testing, \model ~does not require manual labeling effort. 
	\item \model ~develops various mutation strategies to generate the inputs. \model ~leverages unique property of HDC and introduce the distance-guided fuzzing based on the distance between query vector and reference vector to improve fuzzing efficiency. 
	\item Experimental results on MNIST dataset show that \model ~can generate thousands of adversarial inputs with invisible perturbations that can successfully fool the HDC models. On average, \model ~can generate 400 adversarial images within one minute running on commodity computer. We further show that the \model-generated inputs can be used to strengthen the robustness of HDC model. 
\end{itemize}

\section{Related Work}
\label{sec:related}
\mypara{HDC Model}
While previous studies discussed the robustness of HDC with regard to hardware failures such memory errors~\cite{rahimi2016robust}, no known study has systematically look into the algorithmic robustness of HDC. Existing works on HDC focus mainly on two aspects: application of HDC and optimization of HDC processing. For application, Rahimi et al. used HDC on hand gesture recognition and achieve on average 97.8\% accuracy, which surpasses support vector machine by 8.1\%~\cite{rahimi2016hyperdimensional}. HDC was also used for language classification~\cite{rahimi2016robust}. \textit{VoiceHD} applied HDC to voice recognition~\cite{imani2017voicehd}. 
Manabat et al., applied HDC to character recognition and conducted performance analysis~\cite{manabat2019performance}. For optimization of HDC processing, \textit{HDC-IM}~\cite{liu2019hdc} proposed in-memory computing techniques for HDC scenarios based on Resistive RAM. There are also optimizations on HDC targeted at different computing platforms such as FPGA~\cite{schmuck2019hardware} and 3D IC~\cite{wu2018brain}.

\mypara{Fuzz testing}
Fuzz testing has been broadly applied in the software testing community and has shown promising capability for detecting bugs or vulnerabilities in software systems~\cite{fu2019evmfuzzer, chen2019enfuzz, gao2018vulseeker, guo2018dlfuzz}. \textit{American Fuzzy Lop} is a widely used fuzz testing platform that has been applied in industrial systems~\cite{zalewski2014american}. \textit{RFuzz} uses coverage-directed fuzz testing in the hardware testing domain~\cite{laeufer2018rfuzz}. To discover errors occurring only for rare inputs in DNNs, \textit{Tensorfuzz} uses approximate nearest neighbor algorithms and coverage-guided fuzzing (CGF) to enable property-based testing (PBT)~\cite{odena2019tensorfuzz}. Fuzz testing has broader application scenario and high versatility, particularly at systems that are not fully accessible, i.e., grey-box or even black-box systems~\cite{lemieux2018fairfuzz, godefroid2007random}. 

\mypara{Adversarial Generation in DNNs}
Recently, adversarial deep learning~\cite{goodfellow2014explaining, nguyen2015deep}
have demonstrated that state-of-the-art DNN models can be fooled by crafted synthetic images by adding minimal perturbations to an existing image. Goodfellow et al. proposed a fast gradient sign method of generating adversarial examples with 
required gradient computed efficiently using backpropagation~\cite{goodfellow2014explaining}. Nguyen et al. calculated the gradient of the posterior probability for a specific class (e.g., softmax output) with respect to the input image using backpropagation, and then used gradient to increase a chosen unit’s activation to obtain adversarial images~\cite{nguyen2015deep}. While these inspiring methods are effective in generating adversarial examples for DNN, they cannot be applied to HDC because HDC is not built upon solving a mathematical optimization problem to find hyperparameters. The detail of HDC modeling process is described in the next Section. 

\mypara{Our work}
Despite the growing popularity of HDC, there is no known study on systematic HDC testing. This paper presents the first effort in such paradigm by customizing fuzz testing method for HDC models. Further, unique properties of HDC are explored to guide the testing process and the fuzz testing is enhanced with differential testing to enable manual labeling-free testing.

\section{HDC Model Development}
\begin{figure}
	\centering
	\includegraphics[page=1, width=1\columnwidth]{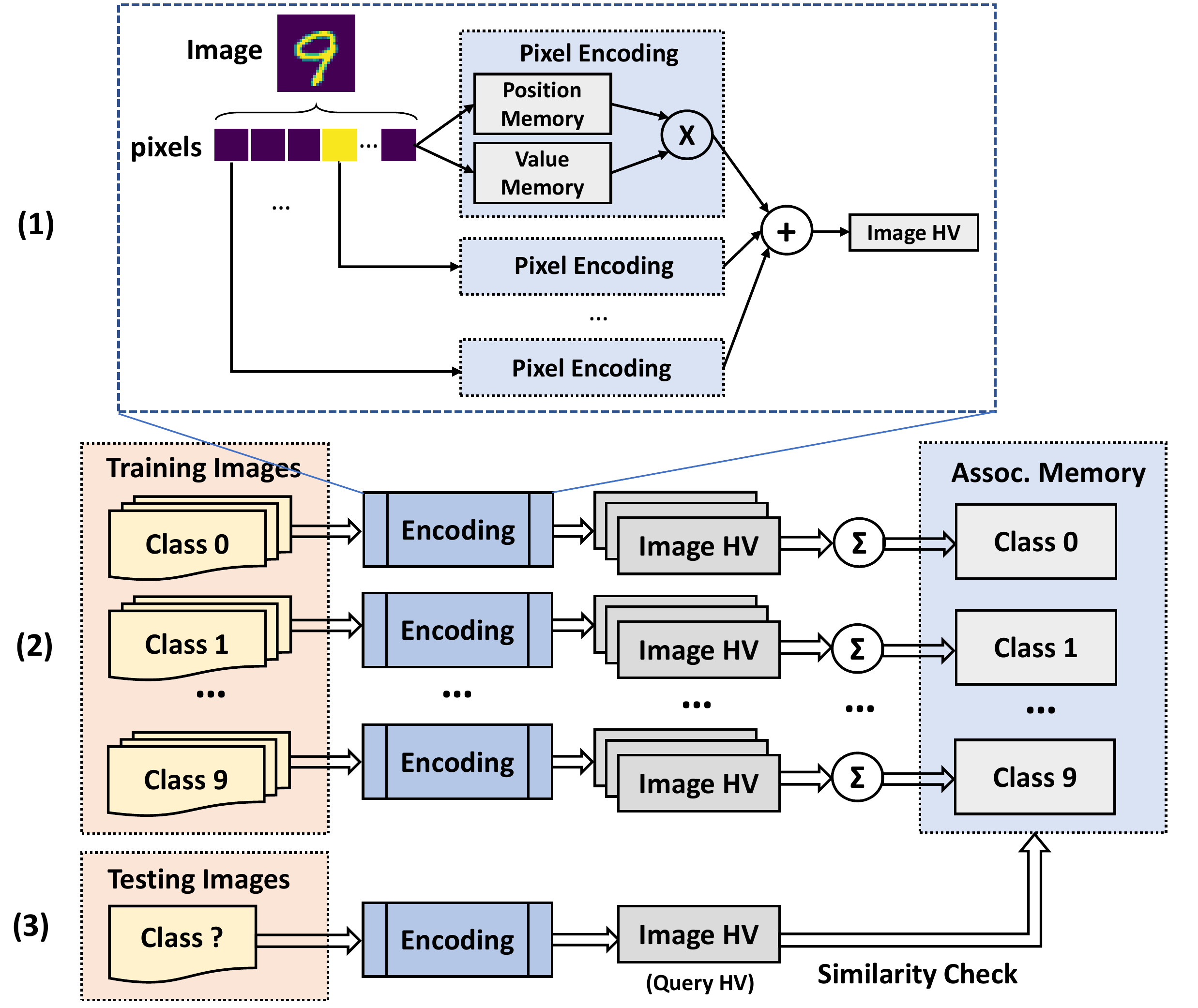}
	\caption{HDC for image classification including: (1) \textbf{Encoding}, (2) \textbf{Training} and (3) \textbf{Testing}. \textbf{Encoding} section maps, or encodes an image into a representative image HV. \textbf{Training} section sums the image HVs from same class to train the Associative Memory (AM). \textbf{Testing} section performs similarity check with the image HV from an unlabelled image with the trained AM to predict its class.}
	\label{fig:hdc}
\end{figure}

We develop an HDC model for image classification as our testing target. We develop three key phases: \textbf{Encoding}, \textbf{Training} and \textbf{Testing} as illustrated in Fig.~\ref{fig:hdc}.

\subsection{Encoding}
The encoding phase is to use HDC arithmetic to encode an image into a hypervector called ``Image HV''. Hypervector (HV) is the fundamental building block of HDC. HVs are high-dimensional, holographic, and (pseudo-)random with independent and identically distributed (i.i.d.) components. In HDC, HV supports three types of arithmetic operation for encoding: addition ($\bigoplus$), multiplication ($\circledast$) and permutation ($\rho$), where ($\bigoplus$) and ($\circledast$) are element-wise addition and multiplication of two operand HVs while ($\rho$) is to perform cyclic shifting of one operand HV. Multiplication and permutation will produce HVs that are orthogonal to the original operand HVs while addition will preserve 50\% of each original operand HVs~\cite{rahimi2016hyperdimensional}.

As shown in Fig.~\ref{fig:hdc} -- Encoding, in order to encode one image into the representing HV, there are three steps. The first step is to decompose and flat the image into an array of pixels. The indices of the pixels in the array reflect the position of the pixel in the original image while the values of the pixels reflect the greyscale level of each pixel. For the MNIST dataset we use in this paper, since the image size is $28 \times 28$ and the pixel range is 0 to 255 in greyscale, we flat a single image into an array with 784 elements with values ranging from 0 to 255. 

The second step is to construct HVs representing each pixel from the index and value information provided by the image array. We randomly generate two memories of HVs: the position HV memory and the value HV memory based on the size and pixel value range of the image. The position HV memory accommodates $28 \times 28 = 784$ HVs, each representing a pixel's position in the original image. The value HV memory accommodates $255$ HVs, each representing a pixel's greyscale value. All the HVs generated are bipolar, i.e, with elements either ``1'' or ``-1''. For each pixel from the image array, we look up the position HV and value HV from the two memories and use multiplication operation to combine them in order to construct the pixel's representing HV. 

The third step is to establish the HV representing the entire image. After constructing all the HVs representing each pixel, the final HV representing the image is established by adding all the pixel HVs together: $ImgHV = Pixel_0HV \bigoplus Pixel_1HV \bigoplus ... \bigoplus Pixel_{783}HV$. So far, we encode one image into a representing image HV that is ready for HDC training and testing. However, addition can destroy the bipolar distribution of the HV (some elements can be numbers other than ``1'' or ``-1''), thus the image HV is bipolarized again by Eq.~\ref{eqn:bipolar}. 

\begin{equation}
    HV[i] =
    \begin{cases} 
      -1 & HV[i] < 0 \\
      1 & HV[i] > 0 \\
      RandomSelect(1, -1) & HV[i] = 0
    \end{cases}
    \label{eqn:bipolar}
\end{equation}

\subsection{Training}
The training phase is to iteratively incorporate the information (or features) contained in each image in the training set into an associative memory (AM) with corresponding label. AM stores a group of HVs, each representing a class. In other words, training is the process of building the AM by adding up all the training images' HVs belong to one class together.
In the beginning, we initialize every element inside the associative memory $AM$ with 0. The dimension of the HVs inside $AM$ is also consistent with the images' HVs. Then, for each image $t_i$ with class label $c_i$ in the training set $T$, we first encode the image into its representing HV $HV_i$ and then we add $HV_i$ into the corresponding class HV inside the associative memory $AM[c_i]$. After one epoch when all the training images are added into $AM$, we finally bipolarize the HVs inside $AM$ again using Eq.\ref{eqn:bipolar} and the $AM$ is constructed and ready for testing and evaluation.

\subsection{Testing}
The testing phase is to evaluate the trained associative memory using the testing dataset. First, for each testing images $t_i$ in the testing set $T$ is encoded into its representing query HV $HV_i$ using the same encoding mechanism in training. We then calculate the cosine similarity $Sim$ between $HV_i$ and every class' hypervector $AM[j]$ in the associative memory: $Sim = Cosim(HV_i, AM[j]) = \frac{HV_i \cdot AM[j]}{||HV_i||||AM[j]||}$. The class with the maximum similarity with $HV_i$ subsequently becomes the prediction result $c_i'$. We then compare the true label $c_i$ with the prediction $c_i'$ for each image to determine if the prediction is correct and to evaluate the accuracy of the HDC model.

\section{HDTest Framework}
\begin{figure}
	\centering
	\includegraphics[page=1, width=0.95\columnwidth]{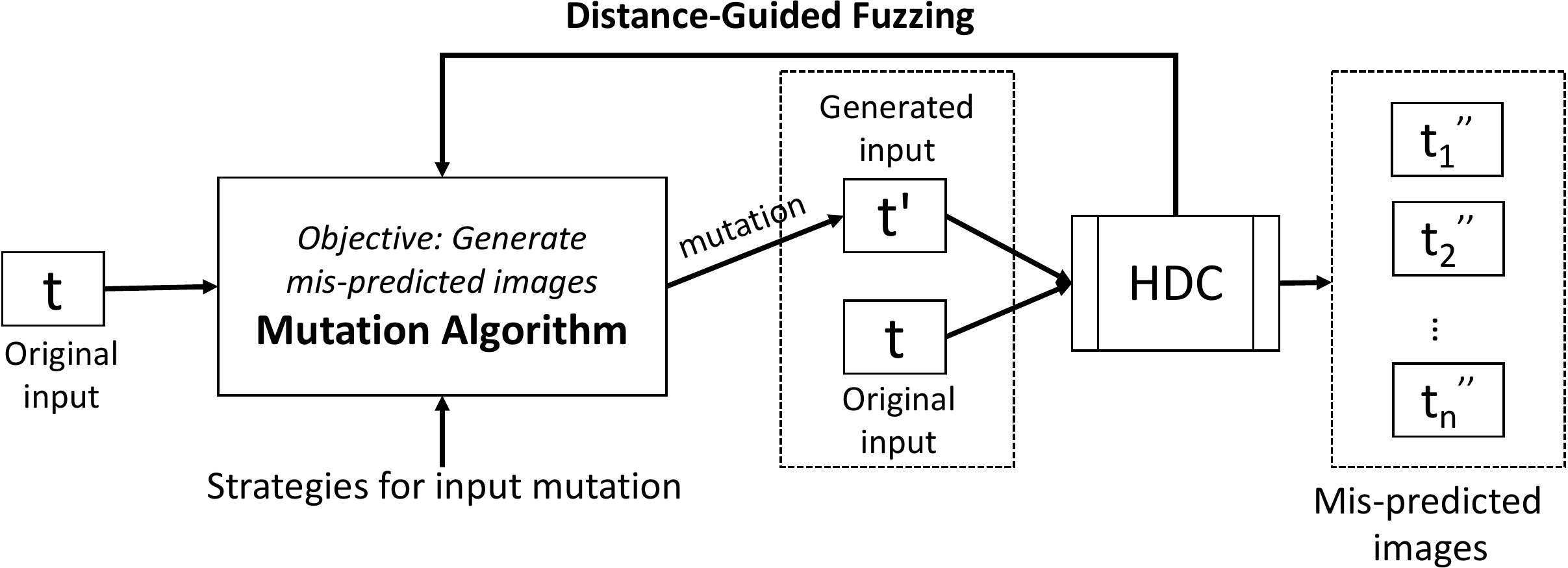}
	\vspace{-0.2cm}
	\caption{Overview of \model}
	\label{fig:arch}
	\vspace{-0.6cm}
\end{figure}

The overview of \model ~is illustrated in Fig.~\ref{fig:arch}. \model ~takes the original input image $t$ without necessarily knowing the label of it. \model ~then applies mutation algorithms on the original input $t$ to generate new input $t'$. Both the generated input and the original input are then sent to the HDC classifier for prediction. We then check if the two predicted labels are different, and if yes, this indicates a successful generation of an adversarial input. Otherwise, \model ~will continue repeating the fuzzing process until success. The details are as following. 

\begin{algorithm}
\small
    \caption{\model ~Mutation Algorithm}
    \algrenewcommand\algorithmicrequire{\textbf{Input}}
    \algrenewcommand\algorithmicensure{\textbf{Output}}
    \begin{algorithmic}[1]
    \Require 
    
    $inputs$: unlabeled input images for testing. 
    
    $HDC$: the HDC model under test.
    
    $strategies$: strategies for mutation listed in Table~\ref{tab:mutate}.
    
    $iter\_times$: maximum allowed iteration times for a seed.
    \Ensure $S$: (set of) adversarial (input) images.
    \State S = []
    \For{$t$ in $inputs$} \\
        \textit{\textbackslash * Prediction of the original input image. */}
        \State $y = HDC(t)$
        \While{$iter\_times$ not reached}
            \State $seeds = strategies(t)$
            \State $y\_seeds = HDC(seeds)$
            \If{$y'$ in $y\_seeds != y$} \\
                \textit{\textbackslash * Indicating successful mis-predcited image generated. */}
                \State $S.append(y')$
                \State $break$
            \Else \\
                \textit{\textbackslash * Continue fuzzing using only the fittest seeds. */}
                \State $seeds = seeds.fittest()$
            \EndIf
        \EndWhile
    \EndFor
    \end{algorithmic}
    \label{alg:fuzz}
\end{algorithm}

The mutation algorithm of \model ~is shown in Alg.~\ref{alg:fuzz}. Its objective is to generate adversarial images by applying mutation strategies to change, or add perturbation on the image. For each input in the unlabelled input image dataset, first, \model ~uses the HDC model to get the predicted label of the original input image (Line 4) as a reference label. Then, \model ~applies mutation strategies on the original image to generate different mutated images as seeds (Line 6). Again, \model ~feeds the seeds into HDC and obtains the corresponding label of the seeds as query labels (Line 7). By comparing the query labels with reference labels, \model ~is able to know if there are any discrepancies which indicate successful generation of an adversarial image. The adversarial images are added to the set $S$ and \model ~proceeds to the next image (Line 10). If all the seeds are still predicted the same as the original input image, \model ~will update the seeds and repeat the process, until a successful adversarial image is generated, or the maximum allowed $iter\_times$ is reached (Line 14).

To maximize the efficiency of the fuzz testing process, \model ~uses distance-guided mutation for seeds update based on the distance between query HV and reference HV. That is, during the mutation process, only the top-N fittest seeds can survive (In our experiments, N = 3). The fitness of seeds are defined as: $fitness = 1 - Cosim(AM[y], HDC(seed))$, where $AM[y]$ is the reference vector in the associative memory and $HDC(seed)$ is the query HV of the seed, encoded by the HDC model. Higher fitness means lower similarity between the HV of the seed and the original input image's HV, indicating higher possibility to generate an adversarial image. Experimental results show that using such guided testing can generate adversarial inputs faster than unguided testing by 12\% on average. 

To ensure the added perturbations are within an ``invisible'' range, we set a threshold for the distance metric during fuzzing (e.g., $L_2 < 1$). When generated images are beyond this limit, it is regarded as unacceptable and then discarded. This constraint can be modified by the user to achieve customized and adaptive performance control when using \model.

We introduce multiple mutation strategies for \model's mutation algorithm as listed in Table~\ref{tab:mutate}. The mutation strategies can be used independently or jointly to implement \model ~with different mutation strategies. In this paper, we focus on four mutation strategies of implementation: \model ~with (1) \textbf{gauss}, (2) \textbf{rand}, (3) \textbf{row \& col\_rand}, and (4) \textbf{shift}. The first two strategies, \textbf{gauss} and \textbf{rand} add Gaussian and random noise onto the image, resulting in holographic mutation on the image. Comparatively, \textbf{row \& col\_rand} only mutates a specific row or column. It is worthwhile to note that \textbf{shift} is different from all the other strategies. \textbf{Shift} does not modify the pixels' values of the image, but just rearranges the pixel locations.

\begin{table}
  \centering
  \caption{Mutation Strategies of \model.}
    \begin{tabular}{cc}
    \toprule
    Name  & Description \\
    \midrule
    row\_rand & randomly mutate all pixels in one single row \\
    col\_rand & randomly mutate all pixels in one single column \\
    rand  & apply random noise over the entire image \\
    gauss & apply gaussian noise over the entire image \\
    shift & apply horizontal or vertical shifting to the image \\
    \bottomrule
    \end{tabular}%
  \label{tab:mutate}%
\end{table}%

\section{Experimental Results}
\label{sec:res}
\subsection{Experimental Setup}
We use the MNIST database of handwritten digits~\cite{lecun1998gradient} as our dataset for training and testing the HDC model at an accuracy around 90\%. 
We use different metrics to evaluate \model: normalized $L_1$ distance and normalized $L_2$ distance between the mutated image and original image, the average fuzzing iterations and the execution time to successfully generate 1000 adversarial images. 
A smaller distance between the generated image and the original image is preferred by \model ~because it means less perturbation added to the image.
The average fuzzing iterations indicate the speed of fuzz testing to obtain adversarial images. One fuzzing iteration refers to a cycle of seed generation, prediction discrepancy check and seed update, as denoted by Line 5 -- 16 in Alg.~\ref{alg:fuzz}. A lower number of iterations means \model ~spends less computation cost and time in generating adversarial images. It is calculated after running \model ~over a set of images, by $Avg. \#iterations = \frac{\#total\_iterations}{\#images}$. To reflect the speed of fuzz testing, we also record the execution time of \model ~on generating 1,000 adversarial images on a commodity computer with AMD Ryzen\texttrademark 5 3600 (6C12T, 3.6GHz) processor with 16GB memory.

\subsection{Mutation Strategies Analysis}
Table~\ref{tab:dist} presents the $L_1$/$L_2$ distance, as well as execution time of generating 1000 adversarial images. Generally, \model ~can generate 1000 adversarial images around 100s - 200s. Among them, \textbf{shift} is the fastest strategy with only 88.4s (because it only changes the pixel locations, or more exactly, indices), while \textbf{rand} is the slowest with 228s, because it requires much higher number of iterations for generation. We provide sample original images, mutated pixels, and generated images of \model ~in Fig. 4--6, under \textbf{gauss}, \textbf{rand}, and \textbf{shift} respectively. 

For the amount of perturbations, \textbf{rand} shows smallest $L_1$ and $L_2$ distance of 0.58 and 0.09, respectively. However, its average fuzzing iterations is 9X higher than that of \textbf{gauss}. Additionally, while \textbf{gauss} shows the lowest average fuzzing iterations of 1.46, its distance metrics are around 5X higher than that of \textbf{rand}.  \textbf{row \& col\_rand} does not perform well, with 9.45 $L_1$ and 0.65 $L_2$ distance and 7.94 average fuzzing iterations, which is outperformed by the \textbf{gauss} in general, therefore we decide to exclude it from further demonstration. 

We can also observe that \textbf{shift} exhibits significantly higher distance and average fuzzing iterations. This however, does not indicate \textbf{shift} has an overall bad performance. Shifting adjusts their geographical location in the image. This can lead to disastrous metrics since all the pixels are mutated and thus we deem the distance metrics are thus not meaningful in reflecting the effectiveness of \textbf{shift} strategy. The average fuzzing iteration of \textbf{shift} is 4.25, which means \model ~on average \textbf{shifts} 4.25 pixels, either horizontally or vertically, to generate an adversarial image.

\begin{figure}
    \centering
    \subfigure[original image]{
        \includegraphics[keepaspectratio, width=.14\columnwidth, page=1]{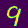}
        \includegraphics[keepaspectratio, width=.14\columnwidth, page=1]{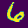}
        \includegraphics[keepaspectratio, width=.14\columnwidth, page=1]{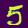}
        \includegraphics[keepaspectratio, width=.14\columnwidth, page=1]{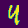}
        \includegraphics[keepaspectratio, width=.14\columnwidth, page=1]{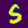}  
    }   
    \subfigure[mutated pixels]{
        \includegraphics[keepaspectratio, width=.14\columnwidth, page=1]{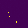}
        \includegraphics[keepaspectratio, width=.14\columnwidth, page=1]{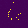}
        \includegraphics[keepaspectratio, width=.14\columnwidth, page=1]{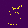}
        \includegraphics[keepaspectratio, width=.14\columnwidth, page=1]{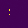}
        \includegraphics[keepaspectratio, width=.14\columnwidth, page=1]{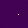}
    }
    \subfigure[generated (adversarial) images]{
        \includegraphics[keepaspectratio, width=.14\columnwidth, page=1]{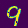}
        \includegraphics[keepaspectratio, width=.14\columnwidth, page=1]{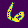}
        \includegraphics[keepaspectratio, width=.14\columnwidth, page=1]{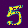}
        \includegraphics[keepaspectratio, width=.14\columnwidth, page=1]{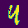}
        \includegraphics[keepaspectratio, width=.14\columnwidth, page=1]{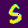}
    }
    \label{fig:case_gauss}
    \caption{Sample adversarial images produced by \model ~with ``gauss''.}
\end{figure}

\begin{figure}
    \centering
    \subfigure[original image]{
        \includegraphics[keepaspectratio, width=.14\columnwidth, page=1]{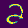}
        \includegraphics[keepaspectratio, width=.14\columnwidth, page=1]{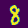}
        \includegraphics[keepaspectratio, width=.14\columnwidth, page=1]{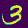}  
        \includegraphics[keepaspectratio, width=.14\columnwidth, page=1]{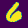}  
        \includegraphics[keepaspectratio, width=.14\columnwidth, page=1]{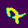}  
    }   
    \subfigure[mutated pixels]{
        \includegraphics[keepaspectratio, width=.14\columnwidth, page=1]{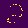}
        \includegraphics[keepaspectratio, width=.14\columnwidth, page=1]{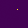}
        \includegraphics[keepaspectratio, width=.14\columnwidth, page=1]{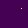}  
        \includegraphics[keepaspectratio, width=.14\columnwidth, page=1]{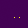}  
        \includegraphics[keepaspectratio, width=.14\columnwidth, page=1]{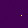} 
    }
    \subfigure[generated (adversarial) images]{
        \includegraphics[keepaspectratio, width=.14\columnwidth, page=1]{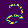}
        \includegraphics[keepaspectratio, width=.14\columnwidth, page=1]{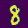}
        \includegraphics[keepaspectratio, width=.14\columnwidth, page=1]{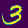}  
        \includegraphics[keepaspectratio, width=.14\columnwidth, page=1]{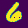}  
        \includegraphics[keepaspectratio, width=.14\columnwidth, page=1]{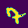} 
    }
    \label{fig:case_rand}
    \caption{Sample adversarial images produced by \model ~with ``rand''.}
\end{figure}

\begin{figure}
    \centering
    \subfigure[original image]{
        \includegraphics[keepaspectratio, width=.14\columnwidth, page=1]{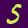}
        \includegraphics[keepaspectratio, width=.14\columnwidth, page=1]{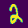}
        \includegraphics[keepaspectratio, width=.14\columnwidth, page=1]{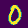}  
        \includegraphics[keepaspectratio, width=.14\columnwidth, page=1]{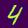}  
        \includegraphics[keepaspectratio, width=.14\columnwidth, page=1]{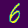} 
    }   

    \subfigure[generated (adversarial) images]{
        \includegraphics[keepaspectratio, width=.14\columnwidth, page=1]{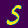}
        \includegraphics[keepaspectratio, width=.14\columnwidth, page=1]{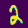}
        \includegraphics[keepaspectratio, width=.14\columnwidth, page=1]{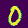}  
        \includegraphics[keepaspectratio, width=.14\columnwidth, page=1]{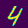}  
        \includegraphics[keepaspectratio, width=.14\columnwidth, page=1]{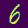} 
    }
    \label{fig:case_shift}
    \caption{Sample adversarial images produced by \model with ``shift''.}
\end{figure}
An interesting observation is that the difficulty of generating adversarial inputs tend to vary for different samples. While some samples require extensive mutations with large perturbations added to obtain an adversarial image such as the third image in Fig. 4 and the first image in Fig. 5, some are easy to generate adversarial images with only few pixel mutations such as the last image in Fig. 4 and the second image in Fig. 5, which we refer to as vulnerable cases. Such vulnerable cases bring potential security loopholes since they can result in incorrect behaviors of HDC models by only minor and even negligible perturbations. Therefore, such images should be emphasized when defending attacks to the HDC systems, and \model ~is able to pinpoint and highlight them.

\begin{table}
  \centering
  \caption{$L_1$ and $L_2$ distance, average fuzzing iterations and runtime of \model ~under different mutation strategies.}
    \begin{tabular}{cccccc}
    \toprule
    \multicolumn{2}{c}{Metric} & gauss & rand  & row \& col\_rand & shift* \\
    \midrule
    \multirow{2}[1]{*}{Avg. Norm. Dist.} & $L_1$    & 2.91  & 0.58  & 9.45  & 10.19* \\
          & $L_2$    & 0.38  & 0.09  & 0.65  & 0.68* \\
    \multicolumn{2}{c}{Avg. \#Iter.} & 1.46  & 12.18 & 7.94  & 4.25 \\
    \multicolumn{2}{c}{Time Per-1K Gen. Img. (s)} & 173.0 & 228.3  & 114.2 &  88.4 \\
    \bottomrule
    \end{tabular}%
  \label{tab:dist}%
\end{table}%

\subsection{Per-class Analysis}
\label{sec:class}
\begin{figure}
	\centering
	\includegraphics[page=1, width=.99\columnwidth]{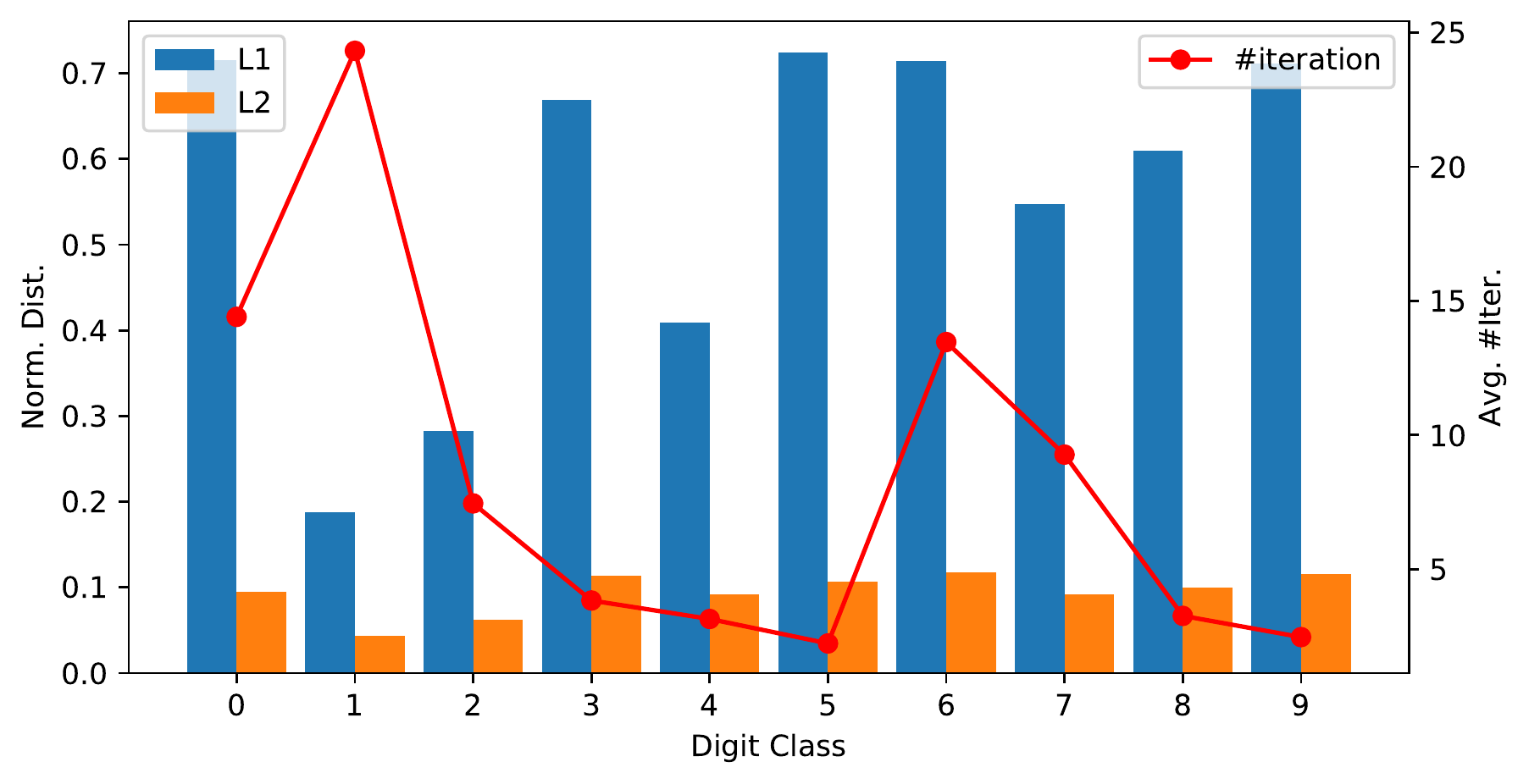}
	\caption{Per-class normalized $L_1$ and $L_2$ distances and average pixel mutations to generate an adversarial image.}
	\label{fig:perclass}
\end{figure}
We then perform a per-class analysis on $L_1$/$L_2$ distance and fuzzing iterations as shown in Fig.~\ref{fig:perclass}. We can observe that for some classes such as ``1'', the average fuzzing iteration is drastically higher, which means that it is relatively difficult to generate adversarial examples. For other classes such as ``9'', the average fuzzing iteration is lower. This is reasonable because all the other digits except for ``7'' are visually dissimilar from ``1'' while ``9'' has quite a few similarities such as ``8'' and ``3''. Moreover, we do not observe apparent relation between fuzzing iteration and distance metrics. For example, the average fuzzing iterations of ``1'' and ``6'' are both high, yet the average distance metrics of ``6'' is noticeably smaller. In summary, smaller average distance and/or lower fuzzing iterations means that \model ~performs better on these digits, or vice versa.

\subsection{Case Study on Defense Against Adversarial Attacks}
\begin{figure}
	\centering
	\includegraphics[page=1, width=.95\columnwidth]{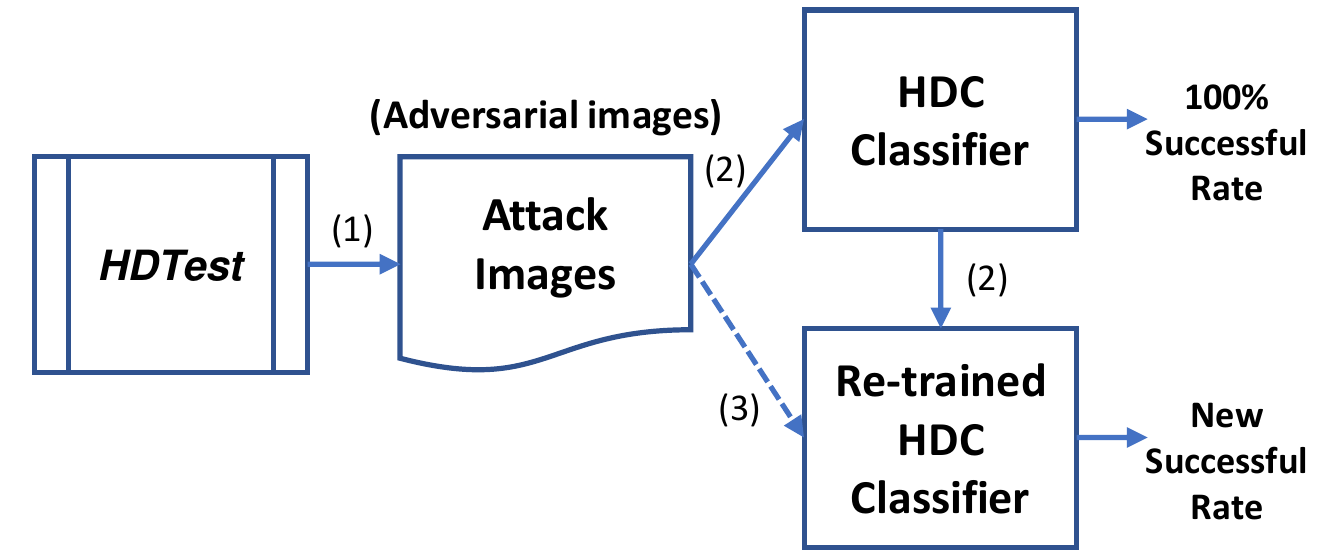}
	\caption{Defend attacks using \model-generated images: (1): Attack image generation; (2): HDC Retraining; (3): Attack HDC again using unseen images.}
	\label{fig:defense}
\end{figure}
Adversarial attack is one major threat to ML systems~\cite{thakur2020evaluating}. The attacker mutates images by adding perturbations and generate adversarial images to fool the model to predict wrong labels. 
As a case study of \model, we propose a defense mechanism to strengthen the robustness of HDC models to such adversarial attacks, as shown in Fig.~\ref{fig:defense}. First, for a specific HDC model, we run \model ~to generate 1000 adversarial images. We then randomly split such 1000 images into two subsets.  
We feed the first subset of adversarial images to the HDC model with correct labels to retrain the HDC model, i.e., updating the reference HVs. Next, we use the second subset of adversarial images to attack the HDC model. Our experimental results show that, after retraining, the rate of successful attack rate drops more than 20\%. 

\subsection{Discussion}
HDC is a fast-growing emerging area that has seen increasing applications in various domains. While optimizing prediction accuracy of existing HDC models is out of scope for this paper, we note intensive ongoing research in this area focusing on advancing HV encoding, model structure, and training mechanism (e.g., retraining)~\cite{nazemi2020synergiclearning}. Complementary and orthogonal to the above efforts, we focus on the problem of how to test HDC models and introduce a customized highly-automated and scalable testing tool for HDC. While we only consider one dataset in this paper due to the fact that HDC is still exploring its use in image classification domain, \model ~can be naturally extended to other HDC model structures because it considers a general greybox assumption with only HV distance information.  

\section{Conclusion}
\label{sec:conclusion}
Brain-inspired HDC is an emerging computational paradigm but is gaining increasing capability in various application domains. This paper presents the first effort in HDC testing by introducing \model, a highly-automated and scalable testing approach without manual labeling. Based on differential fuzz testing, \model ~iteratively mutates inputs to generate adversarial inputs to expose incorrect behaviors of HDC models. We develop multiple mutation strategies in \model, and evaluate \model ~on MNIST dataset. Experimental results show that \model ~is able to generate around 400 adversarial images within one minute. As a case study, we use the \model-generated inputs to retrain HDC models, which can strengthen the robustness of HDC models. While HDC performance still expects more advancements both in theoretical and implementation aspects, this paper aims to raise community attention to the reliability and robustness aspects of this emerging technique.

\end{sloppypar}  
\bibliography{HDFuzz}
\end{document}